\pdfoutput=1

\documentclass[11pt]{article}

\usepackage[]{ACL2023}

\usepackage{times}
\usepackage{latexsym}

\usepackage[T1]{fontenc}

\usepackage[utf8]{inputenc}

\usepackage{microtype}

\usepackage{inconsolata}
\usepackage{bbm}
\usepackage{bm}
\usepackage{graphics}
\usepackage{graphicx}
\usepackage{subfigure}
\usepackage{amsmath}
\usepackage{amssymb}
\usepackage[linewidth=1pt]{mdframed}
\usepackage{times}
\usepackage{latexsym}
\usepackage{booktabs}
\usepackage{multirow}
\usepackage[utf8]{inputenc}
\usepackage{pifont}
\usepackage{soul}
\usepackage{comment}
\title{Mitigating the Learning Bias towards Repetition by Self-Contrastive Training for Open-Ended Generation}

\author{Jian Guan, Minlie Huang\Thanks{~Corresponding author}\\
The CoAI Group, DCST, Institute for Artificial Intelligence, \\
State Key Lab of Intelligent Technology and Systems, \\
Beijing National Research Center for Information Science and Technology, \\
Tsinghua University, Beijing 100084, China \\
\texttt{j-guan19@mails.tsinghua.edu.cn}, \texttt{aihuang@tsinghua.edu.cn}}

\begin{document}
\maketitle
\begin{abstract}
Despite the huge progress in a myriad of generation tasks, pretrained language models~(LMs) such as GPT2 still tend to generate repetitive texts with maximization-based decoding algorithms for open-ended generation. 
We attribute their overestimation of token-level repetition probabilities to the learning bias: LMs capture simple repetitive patterns faster with the MLE loss. We propose self-contrastive training to penalize the output of a premature checkpoint of the same model when it incorrectly predicts repetition, which is shown to mitigate repetition effectively while maintaining fluency on two datasets.
Furthermore, we find that LMs use longer-range dependencies to predict repetitive tokens than non-repetitive ones, which may be 
the cause of sentence-level repetition loops\footnote{The code is available at \url{https://github.com/thu-coai/SelfCont}}. 
\end{abstract}
\section{Introduction}
Existing LMs prefer to generate repetitive texts for open-ended  generation with greedy decoding or beam search~\cite{welleck-etal-2020-consistency}. Even large-scale pretrained LMs such as 
GPT3~\cite{brown2020language} still generate redundant sentences~\cite{dou2022gpt}. Despite many solutions proposed from the perspective of both training~\cite{Welleck2020Neural} and decoding~\cite{holtzman2019curious}, the cause of preference for repetition still needs to be clarified. 

By analyzing the training dynamics of LMs regarding (non-)repetitive tokens, 
we reveal the learning bias towards repetition: LMs capture simple repetitive patterns first, which dominate the output distribution throughout the input space, and then learn more non-repetitive patterns during training. We show that the repetition problem can be mitigated by only training more steps~(i.e., allowing over-fitting), although the coherence with inputs will be impacted. Conversely, when trained insufficiently, LMs will overestimate repetition probabilities even for golden prefixes. 
We propose self-contrastive training~(\textsc{SelfCont}), which exploits the contrast with a premature checkpoint of the same model by penalizing its output when it incorrectly predicts repetition. Experiments on two datasets show that \textsc{SelfCont} effectively alleviates repetition while maintaining fluency by factoring out the undesired repetition behaviors highlighted by the premature checkpoint. 

Besides the above analysis about overestimating token-level repetition probabilities during training,
we also find that 
LMs use longer-range dependencies to predict
repetitive tokens than non-repetitive ones. It may explain why LMs tend to fall into repetition loops~\cite{xu2022learning}. 
The problem may be solved by improving the modeling of long-range dependencies (e.g., increasing model sizes), 
which are left to future work.

\section{Related Work}
Regarding the cause of the repetition problem,
\citet{fu2021theoretical} theoretically derived bounds of repetition probabilities of the first-order Markov LM, although it is difficult to extend the bounds to general LMs. Another line of works attributed repetition to error accumulation during generation 
\cite{Welleck2020Neural,arora2022exposure},
while LMs still prefer repetition given golden prefixes.


We divide recent works that alleviate repetition into training- and decoding-based methods: 
\textbf{(1) Training-based Methods.} \citet{Welleck2020Neural} proposed unlikelihood training~(UL) to reduce the probabilities of repetitive generations. \citet{pmlr-v139-lin21b} and \citet{xu2022learning} further extended the framework at the token and sequence level, respectively. \textsc{SelfCont} focuses on token-level modeling, which is orthogonal with sequence-level methods.
\citet{DBLP:journals/corr/abs-2112-08657} adopted additional modules to learn repetition patterns and control repetition explicitly. 
\textbf{(2) Decoding-based Methods.} 
One straightforward solution to repetition is blocking repetitive $n$-grams generations~\cite{paulus2018deep} or penalizing probabilities of repetitive candidates~\cite{keskar2019ctrl}. \citet{li2022contrastive} selected candidates that maximize the probability difference between different-sized models. Sampling-based decoding methods are also shown effective in avoiding repetition, such as temperature sampling~\cite{ficler2017controlling}, Top-$k$ sampling~\cite{fan2018hierarchical}, nucleus sampling~\cite{holtzman2019curious}, and typical sampling~\cite{meister2022typical}. Although these methods reduce superficial repetition, it is unclear whether they utilize the underlying long-range dependencies to maintain coherence.
\section{Empirical Analysis}

Neural networks (NNs) are highly expressive to approximate arbitrary input-output mappings. Using Fourier analysis, \citet{rahaman2019spectral} showed the
\textit{spectral bias} of NNs: they learn low-frequency components faster during training, which are less complex and vary globally without local fluctuation. 
Our key hypothesis is that simple repetitive patterns may be such low-frequency components and learned by LMs early. In this section, we first formulate LMs~($\S$\ref{lm}), and then investigate the training dynamics~($\S$\ref{ld}) and the ability to model long-range dependencies~($\S$\ref{lrd}) of LMs. 

\subsection{Language Models}\label{lm}
LMs aim to fit the mapping ${x}_t = f({x}_{1:t-1})$ defined by a training corpus, where $x_{1:t}$ is a sequence from the corpus. 
To this end, they are usually trained by minimizing the following cross-entropy loss:
\begin{align}
    \mathcal{L}&=-\textbf{x}_t^\text{T}\cdot\text{log}\big[\text{softmax}\big(f_\theta({x}_{1:t-1})\big)\big],\label{lmloss}
\end{align}
where $\textbf{x}_t\in\{0,1\}^{|\mathcal{V}|}$ is the one-hot representation of $x_t$ indicating its index in the vocabulary $\mathcal{V}$, and $f_\theta({x}_{1:t-1})\in\mathbb{R}^{|\mathcal{V}|}$ is the output logits of the LM parameterized by $\theta$. Predictably, with more training steps, $\text{argmax}(f_\theta$) is closer to the target function $f$. Early stopping~\cite{morgan1989generalization} is a commonly used regularization technique to avoid over-fitting, e.g., stopping training when the validation loss reaches the minimum. Since NNs prioritize learning low-complexity components, early stopping may result in unexpected generations. We are inspired to investigate whether simple repetitive patterns in human-written texts are learned first, thus dominating the generations.



\subsection{Training Dynamics}\label{ld}
We randomly sample 1k sequences containing 512 tokens from the Wikitext-103 dataset~\cite{merity2016pointer} and train GPT2$_{\rm base}$ from scratch for 100 epochs\footnote{We use only 1k samples because we expect to over-fit these samples to observe how repetition in generated texts changes with the fitting degree, considering that it will be very time-consuming to fit the whole Wikitext-103 dataset.}. Given a golden prefix $x_{1:t-1}$, we regard the model prediction $\hat{x}_t=\text{argmax}\big(f_\theta(x_{1:t-1})\big)$ as correct if $\hat{x}_t=x_t$. We call $x_t$ or $\hat{x}_t$ repetitive if it is included in $x_{1:t-1}$, and non-repetitive otherwise.

\begin{figure}[!ht]
\centering
\begin{minipage}[t]{0.48\textwidth}
\centering
\includegraphics[width=\textwidth]{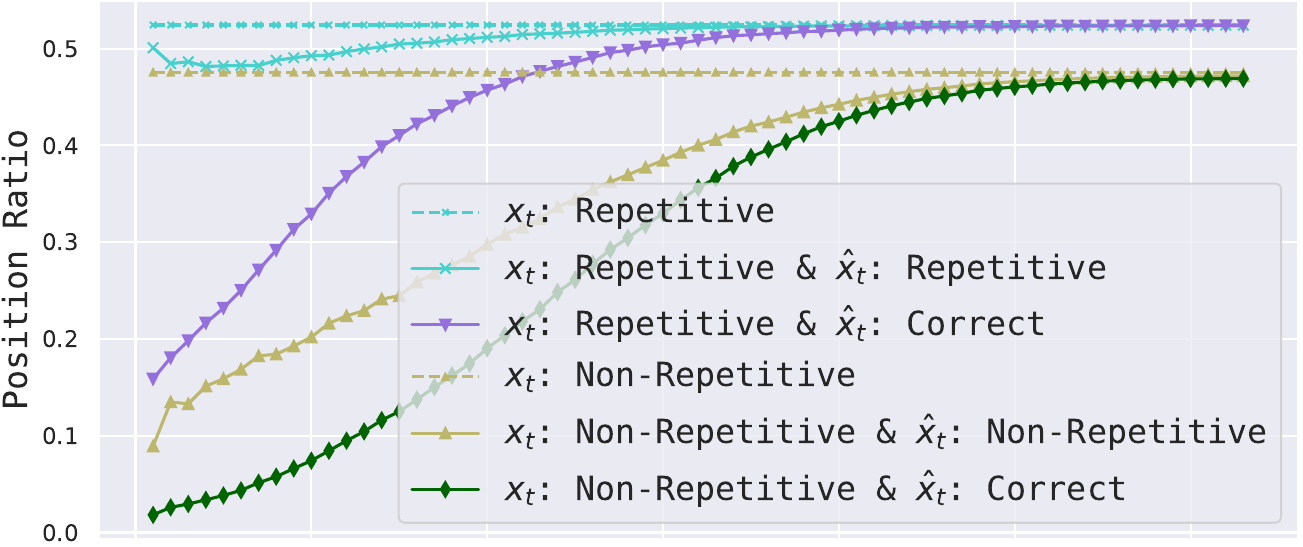}
\end{minipage}
\begin{minipage}[t]{0.48\textwidth}
\centering
\includegraphics[width=\textwidth]{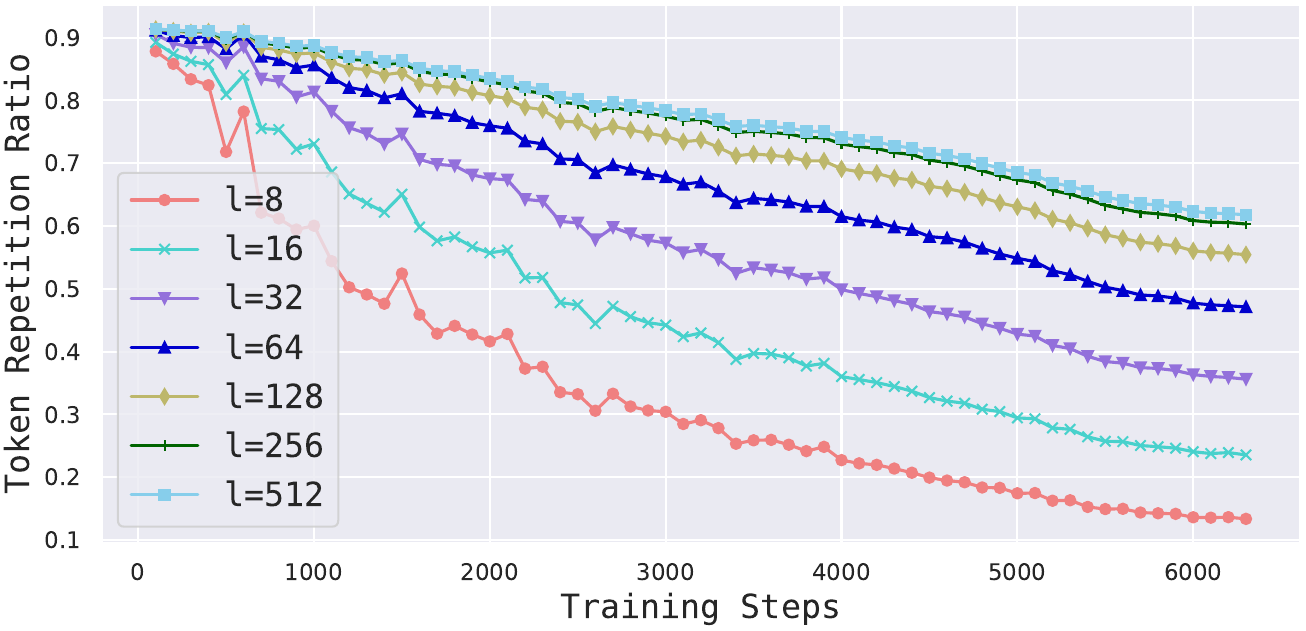}
\end{minipage}
\caption{\textbf{Top}: Ratios of positions where $x_t$ or $\hat{x}_t$ is repetitive 
or not, given golden prefixes of the test set. \textbf{Bottom}: Ratios of tokens that appear in previous $l$ tokens, in model-generated texts with greedy decoding.}
\label{ld_fig}
\end{figure}

\begin{figure*}[!t]
\centering
\begin{minipage}[t]{0.33\textwidth}
\centering
\includegraphics[width=\textwidth]{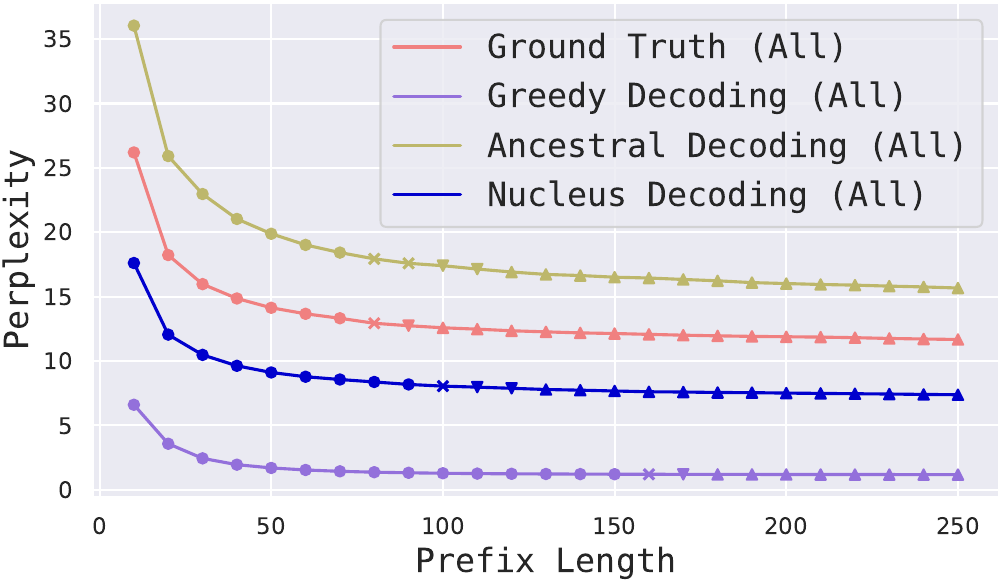}
\end{minipage}
\begin{minipage}[t]{0.315\textwidth}
\centering
\includegraphics[width=\textwidth]{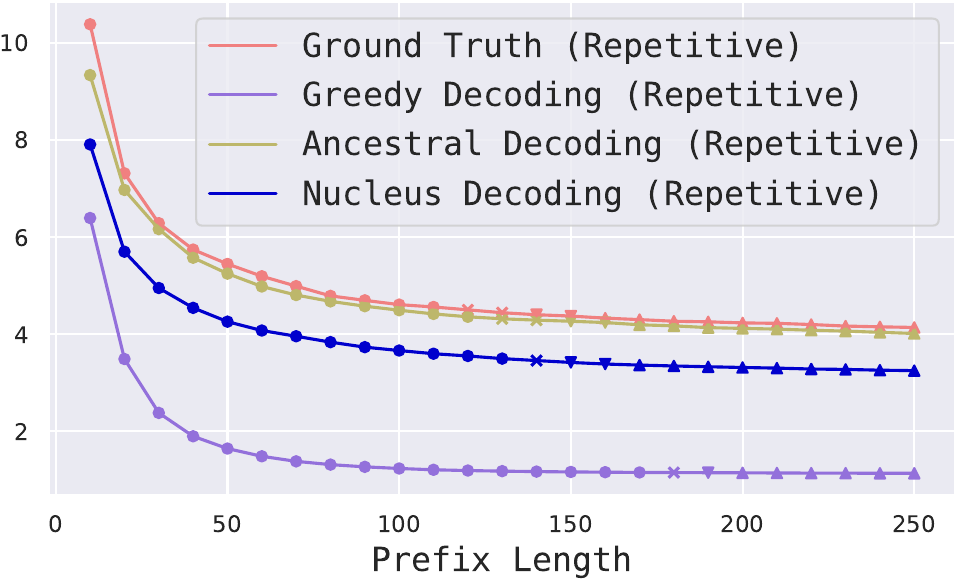}
\end{minipage}
\begin{minipage}[t]{0.315\textwidth}
\centering
\includegraphics[width=\textwidth]{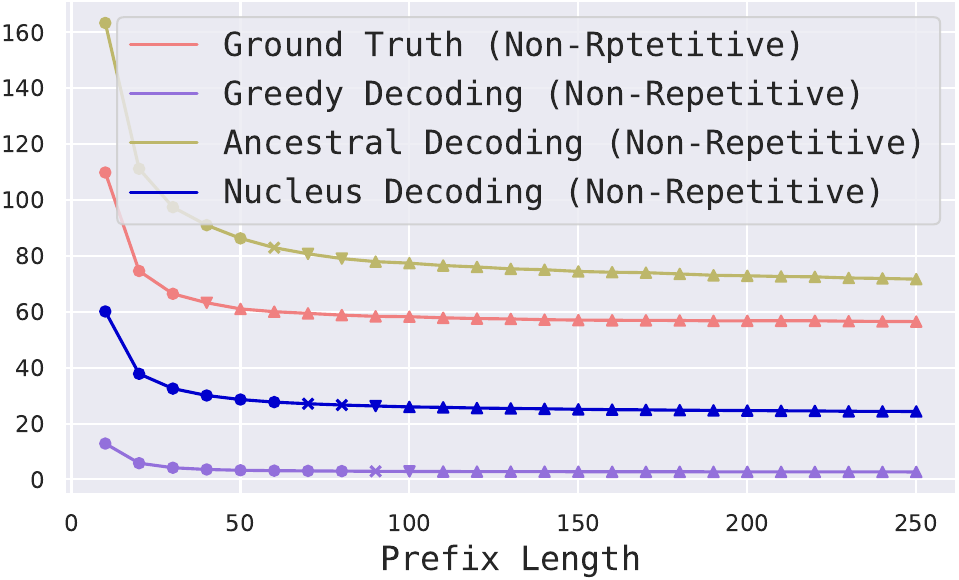}
\end{minipage}
\caption{Perplexity scores computed on \textit{all}, \textit{repetitive} or \textit{non-repetitive} tokens with different prefix lengths. The scores marked with $\bigcirc, \times, \bigtriangledown$ and $\bigtriangleup$ means that the $p$-values compared with the score when the prefix length is 250 fall in the following intervals: $[0, 0.001), [0.001, 0.01), [0.01, 0.05)$ and $[0.05, 1]$, respectively. }
\label{ppl}
\end{figure*}

Figure~\ref{ld_fig} plots the training curves, revealing the learning bias of the LM: (1) The initially learned components prefer to copy input tokens throughout the input space, as indicated by predicting repetitive tokens at $\sim$90\% of positions for both golden and generated prefixes. 
(2) With golden prefixes, at those positions where $x_t$ is repetitive, the LM almost always predicts repetition during training. When $x_t$ is non-repetitive, the LM predicts more non-repetitive tokens with more training steps. The repetition ratio also gradually decreases in model-generated texts.  
(3) The token prediction accuracy improves faster when $x_t$ is repetitive, indicating that the LM learns repetitive patterns more easily. 
Moreover, we notice that the validation loss rises at the 1,500th step, where the LM predicts much more repetitive tokens than the ground truth. At the end of the training, the generation has a closer token repetition ratio to the ground truth. But manual inspection finds the coherence with inputs is poor due to over-fitting.
Appendix~\ref{app1} shows several generation cases. 

\subsection{Modeling Long-Range Dependencies}\label{lrd}
Figure~\ref{ld_fig} (Top) shows that LMs are still able to predict non-repetitive tokens conditioned on golden prefixes. However, it is still unclear why they get into repetition loops during generation and do not generate any non-repetitive tokens.
To shed light on this behavior, we further investigate how LMs learn and utilize long-range dependencies. We fine-tune GPT2$_{\rm base}$ on the training set of Wikitext-103, and examine the effect of prefix lengths on the perplexity of tokens that have appeared in the previous 250 tokens~(called \textit{repetitive}) or not on the original test set and model-generated texts. 

Figure~\ref{ppl} indicates \textbf{(1) The LM only learns 
dependencies within $\sim$100 tokens overall.} When the prefix length is larger than 100, the perplexity on golden tokens no longer drops significantly ($p\geqslant0.05$). 
\textbf{(2) The LM learns and utilizes longer-range dependencies to predict repetitive tokens than non-repetitive ones.} 
The perplexity on golden repetitive/non-repetitive tokens plateaus 
when the prefix length is larger than 160/50, respectively. The case is similar for generated texts. \textbf{(3) The LM uses short-range contexts to predict non-repetitive tokens regardless of decoding algorithms.} Contexts beyond 100 tokens hardly help predict non-repetitive tokens, 
implying sampling-based decoding reduces repetition through randomness instead of using long-range dependencies.

Based on the above observation, we conjecture that the LMs keep repeating the same sentence with maximization-based decoding~\cite{xu2022learning} because they rarely learn long-range non-repetitive patterns beyond the sentence level. When generating long texts, LMs may struggle to maintain non-repetitive within a long range.
To test the idea, we train GPT2$_{\rm base}$ from scratch on three datasets constructed from the training set of Wikitext-103: (1) $\mathcal{D}_{\rm original}$, where examples are directly sampled from the original training set; 
(2) $\mathcal{D}_{\rm random}$, where each example contains 30 randomly sampled sentences; (3) $\mathcal{D}_{\rm norept}$, where each example also contains 30 random sentences, but there is at most one token overlapping  
between any adjacent 5 sentences~(generally the period ``.''). 
Each dataset consists of 20k examples. 
We then generate texts using greedy decoding conditioned on the first 50 tokens in the original test set and compute the ratio of texts which fall into loops~\cite{holtzman2019curious}.

\begin{table}[!ht]
\small
    \centering
    \begin{tabular}{c|ccc}
    \toprule
    \textbf{Training Sets}&$\mathcal{D}_{\rm original}$&$\mathcal{D}_{\rm random}$&$\mathcal{D}_{\rm norept}$\\
    \midrule
    \textbf{Ratios (\%) $\downarrow$}&60.42&96.04&1.67\\
    \bottomrule
    \end{tabular}
    \caption{Ratios of texts which get stuck into loops generated by LMs trained on different training sets.}
    \label{reprep}
\end{table}

As shown in Table~\ref{reprep}, compared to $\mathcal{D}_{\rm original}$, the LM trained on $\mathcal{D}_{\rm random}$ has higher repetition ratios because it learns shorter-range non-repetitive patterns only within one sentence. Besides, although sentences in each $\mathcal{D}_{\rm random}$  example are unrelated, they can contain repetitive tokens\footnote{The ratios of tokens that have appeared in previous 128 tokens are 12.52\% and 32.05\%
for the training sets of $\mathcal{D}_{\rm original}$ and $\mathcal{D}_{\rm random}$,
respectively. $\mathcal{D}_{\rm random}$ has even more repetition than $\mathcal{D}_{\rm original}$ possibly because random sentences  repeat high-frequency words than human-written sentences.}, making the LM learn spurious long-range repetitive patterns to get into repetition loops. In contrast, the LM trained on $\mathcal{D}_{\rm norept}$ rarely gets into loops since it learns both repetitive and non-repetitive patterns almost within one sentence. Specifically, any adjacent five sentences in each $\mathcal{D}_{\rm norept}$ example are unrelated and hardly share tokens. These findings empirically support our hypothesis. Appendix~\ref{app2} shows more details.

\begin{table*}[!ht]
\tiny
    \centering
    \begin{tabular}{l||cc|ccc|cc||cc|ccc|cc}
    \toprule 
\textbf{Models}&\textbf{PPL}&\textbf{MAUVE}&\textbf{R-16}&\textbf{R-32}&\textbf{R-128}&\textbf{D-3}&\textbf{D-4}&\textbf{PPL}&\textbf{MAUVE}&\textbf{R-16}&\textbf{R-32}&\textbf{R-128}&\textbf{D-3}&\textbf{D-4}\\
    \midrule
    \midrule
    \textit{Greedy}&\multicolumn{7}{c||}{\textit{Dataset: \textit{Wikitext-103}}}&\multicolumn{7}{c}{\textit{Dataset: \textit{WritingPrompts}}}\\
    \midrule
    \textbf{MLE}&2.55&3.29&41.23&70.18&83.28&19.27&23.95&1.76&0.61&71.08&87.20&89.43&9.61&11.40\\
    \textbf{UL}&3.20&7.16&33.91&61.90&76.89&25.13&31.90&2.01&1.63&59.43&81.63&85.89&11.66&14.30\\
    \textbf{ScaleGrad}&\underline{4.61}&\underline{7.66}&\underline{29.82}&\underline{50.69}&\underline{66.14}&\underline{36.96}&\underline{47.34}&\underline{2.87}&\underline{11.17}&\underline{52.29}&\underline{69.53}&\underline{76.16}&\underline{18.16}&\underline{24.40}\\
    \midrule
    \textbf{\textsc{SelfCont}}&\textbf{6.47}&\textbf{17.34}&\textbf{23.29}&\textbf{39.41}&\textbf{62.46}&\textbf{46.71}&\textbf{57.66}&\textbf{3.30}&\textbf{20.05}&\textbf{35.13}&\textbf{53.69}&\textbf{74.09}&\textbf{23.30}&\textbf{31.52}\\
    \midrule
    \midrule
    \textit{Nucleus}&\multicolumn{7}{c||}{\textit{Dataset: \textit{Wikitext-103}}}&\multicolumn{7}{c}{\textit{Dataset: \textit{WritingPrompts}}}\\
    \midrule
    \textbf{MLE}&\underline{20.66}&21.09&19.40&30.22&48.11&\underline{71.92}&\textbf{84.75}&{18.68}&\underline{88.54}&20.95&32.53&48.87&60.38&81.55\\
    \textbf{UL}&15.54&21.78&\underline{18.45}&29.57&\textbf{46.69}&69.63&82.87&\underline{19.39}&81.49&\underline{18.36}&27.98&\textbf{42.65}&\textbf{63.92}&\underline{82.93} \\
    \textbf{ScaleGrad}&12.41&\underline{25.69}&18.59&\textbf{29.24}&\underline{45.19}&66.35&80.23&14.14&77.82&18.62&\underline{27.80}&\underline{41.22}&56.74&77.27\\
    \midrule
    \textbf{\textsc{SelfCont}}&\textbf{19.02}&\textbf{34.37}&\textbf{16.45}&\underline{26.47}&45.10&\textbf{72.02}&\underline{84.78}&\textbf{19.86}&\textbf{89.84}&\textbf{17.56}&\textbf{26.98}&43.39&\underline{63.33}&\textbf{83.51}\\
    \midrule
    \midrule
    \textbf{Ground Truth}&\textit{18.31}&\textit{100}&\textit{17.38}&\textit{27.92}&\textit{46.29}&\textit{72.34}&\textit{84.20}&\textit{24.01}&\textit{100}&\textit{16.36}&\textit{26.47}&\textit{42.30}&\textit{74.49}&\textit{90.01}\\
    \bottomrule
    \end{tabular}
    \caption{Automatic evaluation results with greedy and nucleus decoding on Wikitext-103 and WritingPrompts.}
    \label{rep_result}
\end{table*}

\section{Self-Contrastive Training}

We denote the premature checkpoint  as $f_{\theta_0}$, 
which frequently predicts repetitive tokens. 
Formally, the \textsc{SelfCont} algorithm is formulated 
as follows:
\begin{align}
    f_{\theta}&=f_{\theta_1}+\text{sg}(wf_{\theta_0}),\\
    w&=\lambda\mathbbm{1}({x_t\not\in x_{1:t-1}})\mathbbm{1}({\hat{x}_t\in x_{1:t-1}})\label{wfunc}\\
\hat{x}_t&=\text{argmax}\big(f_{\theta_0}(x_{1:t-1})\big),
\end{align}
where $\text{sg}(\cdot)$ means stopping back-propagation of gradients, $\lambda$ is a tunable hyper-parameter to control the extent of repetition penalty, and $\mathbbm{1}$ is the indicator function. $f_{\theta_1}$ is the target LM initialized from $f_{\theta_0}$, and we 
optimize $f_\theta$ using Eq.~\ref{lmloss} until the validation loss converges to the minimum. 
The gradient for each token $u\in\mathcal{V}$ has changed to: 
\begin{align} \nabla_{u}\mathcal{L}=&\frac{\text{exp}(f_{\theta_1}|_{u})}{\sum_{v\in\mathcal{V}}w_{v,u}\text{exp}(f_{\theta_1}|_{v})}-\mathbbm{1}({u=x_t}),\\
w_{v,u}=&\text{exp}\big(w(f_{\theta_0}|_{v}-f_{\theta_0}|_u)\big),
\end{align}
where $f_{\theta_1}|_u$ is the output of $f_{\theta_1}$ at the $u$-th dimension. 
If $w$ is $0$, $w_{v,u}$ is always $1$ and $\nabla_{u}\mathcal{L}$ degenerates to the same as the vanilla LM. If $w$ is not $0$ and $u$ is not $x_t$, tokens with high logits under $f_{\theta_0}$ will receive larger gradients than the vanilla LM since $w_{v,u}$  is mostly smaller than $1$ with different $v$. 
As for $u=x_t$~($w\not=0$), it may also be penalized with a positive gradient if $f_{\theta_0}|_u$ is large enough, which usually means a dull token.
By penalizing components that excessively prefer repetitive or dull tokens highlighted by $f_{\theta_0}$,
$f_{\theta_1}$ can utilize more complex patterns learned later to generate texts. 

\section{Experiments}
\paragraph{Datasets} 
We conduct experiments on Wikitext-103~\cite{merity2016pointer} and WritingPrompts~\cite{fan2018hierarchical}. The prompt and story in each WritingPrompts example are concatenated as a sequence. We set the maximum sequence length to 512 and take the first 50 tokens as input to generate the rest. Table \ref{stat} presents the detailed statistics.

\begin{table}[!ht]
\scriptsize
    \centering
    \begin{tabular}{l|cccc}
    \toprule
    \textbf{Datasets}&\textbf{|Train|}&\textbf{|Validation|}&\textbf{|Test|}&\textbf{Avg. Len}\\
    \midrule
    \textbf{Wikitext-103}&201,632&448&480&512\\    \textbf{WritingPrompts}&272,600&15,620&15,138&439\\
    \bottomrule
    \end{tabular}
    \caption{Statistics of the datasets.}
    \label{stat}
\end{table}
\paragraph{Baselines} We compare \textsc{SelfCont} to three baselines: MLE, token-level UL~\cite{Welleck2020Neural} and ScaleGrad~\cite{pmlr-v139-lin21b}. Since \textsc{SelfCont} focuses on token-level modeling, we do not compare it to sentence-level methods that 
directly penalize repetition loops, e.g., DITTO~\cite{xu2022learning}. 

\paragraph{Implementation} All baselines are implemented based on GPT2$_{\rm base}$. We set the batch size to 16, the learning rate to 1e-4, and $\lambda$ in Eq.~\ref{wfunc} to $4.0$. For \textsc{SelfCont}, we fine-tune GPT2$_{\rm base}$ for one epoch using MLE and take the checkpoint as $f_{\theta_0}$ for both datasets. 
We use different $p$ for different models based on the performance on the validation set. Appendix~\ref{hyperparam} shows more details.

\paragraph{Metrics}  
We use perplexity (PPL) under GPT2$_{\rm xl}$ to evaluate fluency, MAUVE~\cite{mauve} to measure the similarity between golden and generated distributions, the token repetition ratios~(R-$l$) to measure the ratio of tokens that appear in previous $l$ tokens~\cite{Welleck2020Neural}, and distinct~(D-$n$)~\cite{DBLP:conf/naacl/LiGBGD16} to evaluate the $n$-gram diversity. The closer scores to the ground truth mean better quality for all metrics. 

\paragraph{Results} As shown in Table~\ref{rep_result}, \textsc{SelfCont} outperforms baselines in all metrics using greedy decoding. However, the high R-128 score shows it can still generate repetition loops due to the disability of small-scale LMs to model long-range dependencies. Using nucleus decoding, we see that different baselines can achieve similar repetition ratios and diversity to the truth by tuning $p$, while \textsc{SelfCont} has better fluency and higher MAUVE scores. 

\section{Conclusion}
We present empirical studies on LMs' preference for repetition by analyzing the training dynamics, which highlights their learning bias towards simple repetitive patterns. We propose penalizing outputs of a premature checkpoint during training, which effectively mitigates repetition while maintaining fluency. We also provide insight into why LMs easily fall into repetition loops by showing their disability to model long-range dependencies. 
Sampling-based decoding reduces repetition through randomness but not utilizing long-range dependencies. We believe that maximization-based decoding can also generate coherent texts without repetition by improving the modeling of long-range dependencies, which is left to future work.
\section*{Acknowledgments}
This work was supported by the National Science Foundation for Distinguished Young Scholars (with No. 62125604) and the NSFC projects (Key project with No. 61936010). This work was also supported by the Guoqiang Institute of Tsinghua University, with Grant No. 2020GQG0005.

\section{Limitations}
The limitations of this paper mainly lie in the following folds: \textbf{(1)} We do not provide any theoretical analysis for the correlation between long-range dependencies and repetition loops, as well as solutions to avoid repetition loops with maximization-based decoding. \textbf{(2)} We do not discuss the source of LMs' learning bias, which may be caused by multiple factors, such as the Transformer architecture~\cite{vaswani2017attention}, the MLE loss, or the auto-regressive generation manner. \textbf{(3)} We conduct experiments based on GPT2 due to resource limitations. The conclusions may differ for extra-large LMs~(such as GPT3). \textbf{(4)} We do not experiment with RNN-based models, which are also shown to prefer repetition~\cite{elman1990finding}. 
\textbf{(5)} We do not perform the manual evaluation to compare \textsc{SelfCont} with baselines since we focus on repetition in this paper, which can be automatically evaluated reliably. Perplexity and mauve scores are also shown to correlate highly with manual evaluation for evaluating fluency and overall quality, respectively.





\bibliography{anthology,custom}
\bibliographystyle{acl_natbib}

\appendix\label{sec:appendix}

\section{Details for Empirical Analysis}
\subsection{Training Dynamics}\label{app1} Table~\ref{case_overfit} shows several cases generated by the LM with greedy decoding at different training steps. We summarize the findings as follows: \textbf{(1)} In the beginning, the LM keeps repeating the high-frequency word ``<eos>,'' indicating that it does not capture phrase-level dependencies yet. \textbf{(2)} At the 1500th step, the LM first generates a few fluent sentences and then gets stuck into the repetition of ``the building,'' showing that it learns long-range dependencies conditioned on the golden prefix while the repetitive patterns dominate the probability distributions conditioned on the generated prefix. This case suggests the global tendency towards repetition for out-of-distribution inputs. \textbf{(3)}
At the 6000th step, the LM can generate long, fluent texts without repetition. However, it is difficult for the LM to maintain coherence with inputs due to over-fitting. For example, in the generated first sentence, ``she had begun in 1962,'' ``she'' conflicts with ``he'' in the input.

\begin{table*}[!ht]
\tiny
    \centering
    \begin{tabular}{l|p{390pt}}
    \toprule 
    \textbf{Training Steps}&\textbf{Cases}\\
    \midrule
    \textbf{100}&\textbf{<eos> = Robert Boulter = <eos> <eos> Robert Boulter is an English film , television and theatre actor . He had a guest @-@ starring role on the television series The Bill in 2000 .} The first , and the " . The " The " <eos> <eos> <eos> <eos> <eos> <eos> <eos> <eos> <eos> <eos> <eos> <eos> <eos> <eos> <eos> <eos> <eos> <eos> <eos> <eos> <eos> <eos> <eos> <eos> = = = = = = = = = = = = = = = = = = = = = = = = = = = = = = = = = = = = = = = = = = = = <eos> <eos> <eos> <eos> <eos> <eos> <eos> <eos> <eos> <eos> <eos> <eos> <eos> <eos> <eos> <eos> <eos> <eos> <eos> <eos> <eos> <eos> <eos> <eos> <eos> <eos> <eos> <eos> <eos> <eos> <eos> <eos> <eos> <eos> <eos> <eos> <eos> <eos> <eos> <eos> <eos> <eos> <eos> <eos> <eos> <eos> <eos> <eos> <eos> <eos> <eos> <eos> <eos> <eos> <eos> <eos> <eos> <eos> <eos> <eos> <eos> <eos> <eos> <eos> <eos> <eos> <eos> <eos> <eos> <eos> <eos> <eos> <eos> <eos> <eos> <eos> <eos>\\
    \midrule
    \textbf{1500}&\textbf{<eos> = Robert Boulter = <eos> <eos> Robert Boulter is an English film , television and theatre actor . He had a guest @-@ starring role on the television series The Bill in 2000 .} It was a new project in November 18th century , after a new new building to the building , was created by the new new building . It was built in 18th century . The building of the new building , was built in 1966 , which had been created a new building . The building , was built in the new building . In the building of the building , which was built in the building by the building by new building by the building . The building , the new building by new building by new building , the building , and , and work , in the building . <eos> In 2009 , the building , the building , the building , the building . The building , the building , the building . The building by the building , the building , the building , the building by the building , and new building , the building , the building , which included by the building . <eos> <eos> <eos> = = = = = = <eos> <eos> The building , the building , the building . The building is the building , the building , the building , the building , the building and building . The building and building , the building , the building , in the building , the building , the building and building and building , the building , the building . The building and building , the building , the building , the building , the building , the building , the building . In the building , the building , the building and building , the building . In the building , the building , the building , the building , the building and building and building and building , the building , the building , the building , the building and building , the building , the building , the building , the building . <eos> <eos> <eos> <eos> The building , the building , the building , the building , the building , the building , the building , the building , the building and building , the building , the building , the building , the building , the building , the building and building , the building , the building and building , the building , the building , the building . <eos> The building and the building and building , the building , and , the building and building , the building , the building , the building ,\\
    \midrule
    \textbf{6000}&\textbf{<eos> = Robert Boulter = <eos> <eos> Robert Boulter is an English film , television and theatre actor . He had a guest @-@ starring role on the television series The Bill in 2000 .} She had begun in 1962 , was built a series of New York Times in 1962 – 1938 after producing the Laidlay , and described as well as well as The Lastunk> 's assistant ... ] named " . She later explained : " [ 500 people who did not be turned off for me to me to me . She decided to his own song " As Rocham R. Nails , saying " . She knew it 's script was going to make me to live film , and I mean it was through me to get it 't know what we 't know what we want to make me . <eos> On 17 November 1953 , and it was nominated for me . He had done what he wanted the BBC at a period at her following year because it appearance at the mid @-@ selling artist with her singing . She was nominated for the background scene in 1962 . She spent four days after getting part of the public appearances at the war appearance , California , and ] base . " <eos> = = <eos> <eos> Aftermath Meodie Glam artist of the summer May 1967 , New York Times , and the war as the Star Wars franchise . He began to use the National Association ( from his staff ; it was included Star ) , including the Lyds house east , and I was the West Virginia Tech back to the war and Mennon ; there . She developed by <eos> William Peninsular League <eos> = = <eos> The script , 2004 . The script was named after the North America for the LAM passed . The script was commissioned to the American co @-@ person to produce producer ( present , taking place of the mid @-@ old , and the mid @-@ old @-@ old film , The Next Generation . The New York Times , having won the 4th birthday of the 4 , in the 4 million viewers . This was announced that it was cut of the media . The Elder Scrolls IV of the production , in East Coaster and The company entitled The Next Generation . <eos> For example , including the war , having performed on 6 , having released in East Coast Division . <eos> Upon its crew became a series of the produce\\
    \bottomrule
    \end{tabular}
    \caption{Generation cases with greedy decoding at different training steps to investigate the training dynamics. The inputs are highlighted in \textbf{bold}.}
    \label{case_overfit}
\end{table*}
\subsection{Long-Range Dependencies}\label{app2}
\begin{figure*}[!t]
\centering
\begin{minipage}[t]{0.33\textwidth}
\centering
\includegraphics[width=\textwidth]{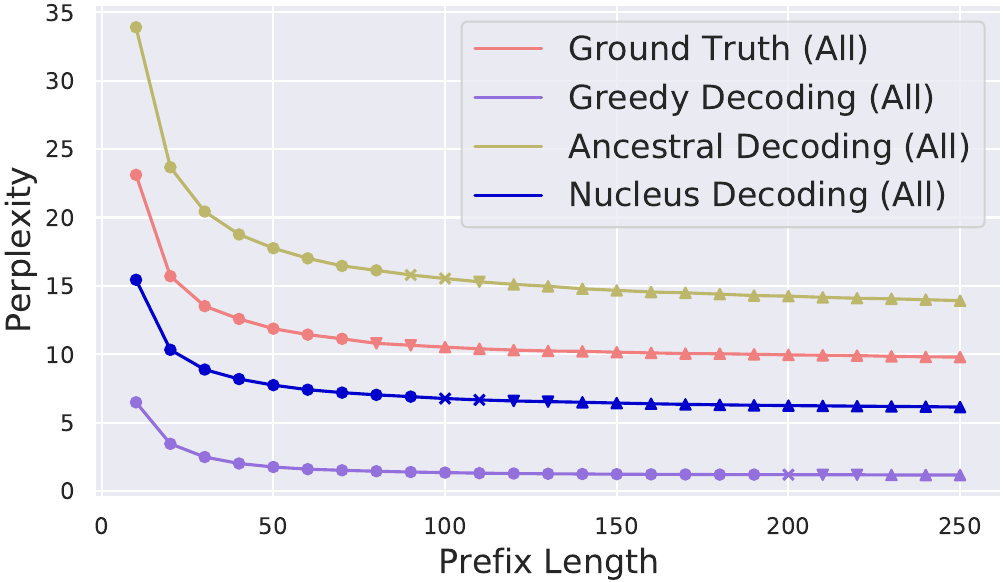}
\end{minipage}
\begin{minipage}[t]{0.315\textwidth}
\centering
\includegraphics[width=\textwidth]{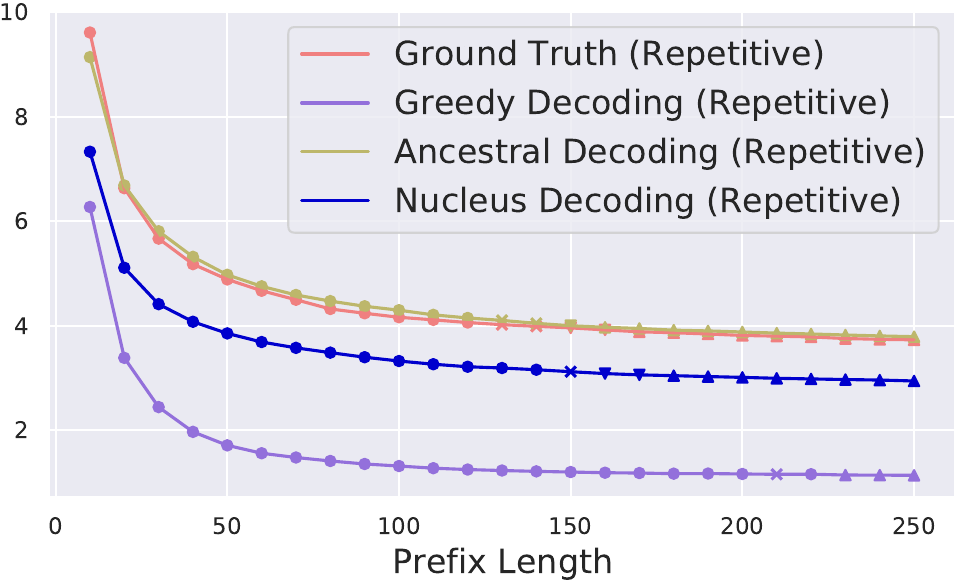}
\end{minipage}
\begin{minipage}[t]{0.315\textwidth}
\centering
\includegraphics[width=\textwidth]{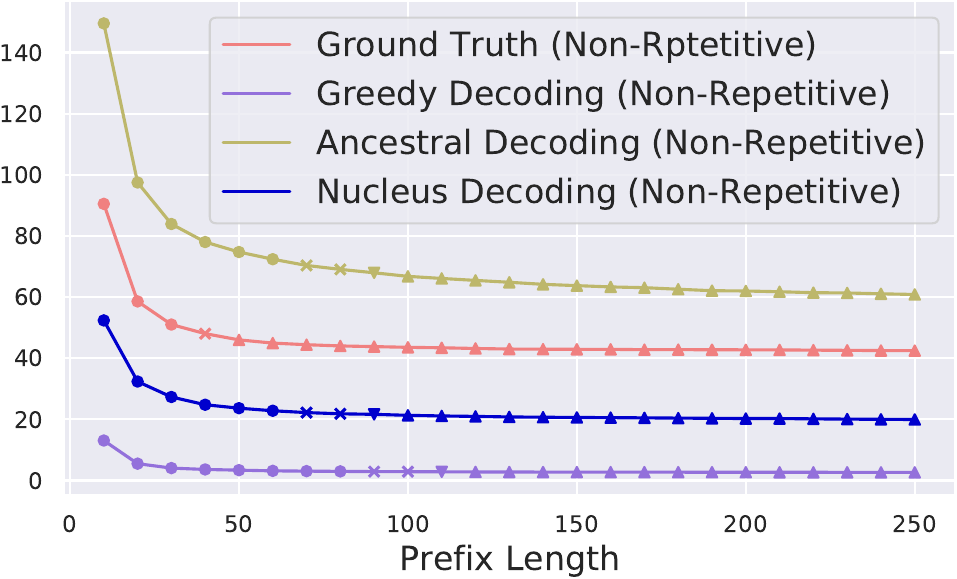}
\end{minipage}
\caption{Perplexity scores computed on \textit{all}, \textit{repetitive} or \textit{non-repetitive} tokens with different prefix lengths based on GPT2$_{\rm large}$. The scores marked with $\bigcirc, \times, \bigtriangledown$ and $\bigtriangleup$ means that the $p$-values compared with the score when the prefix length is 250 fall in the following intervals: $[0, 0.001), [0.001, 0.01), [0.01, 0.05)$ and $[0.05, 1]$, respectively. }
\label{ppl_large}
\end{figure*}

\paragraph{Observation} For the experiment in Figure~\ref{ppl}, we generate texts with three decoding algorithms conditioned on the first 50 tokens on the test set. Ancestral decoding means directly sampling tokens from the original probability distribution. For nucleus decoding, we set $p$ to 0.9. Figure~\ref{ppl_large} shows the performance of GPT2$_{\rm large}$, which shows similar results with GPT2$_{\rm base}$ in Figure~\ref{ppl}.

\begin{figure}[!h]
\centering
\includegraphics[width=\linewidth]{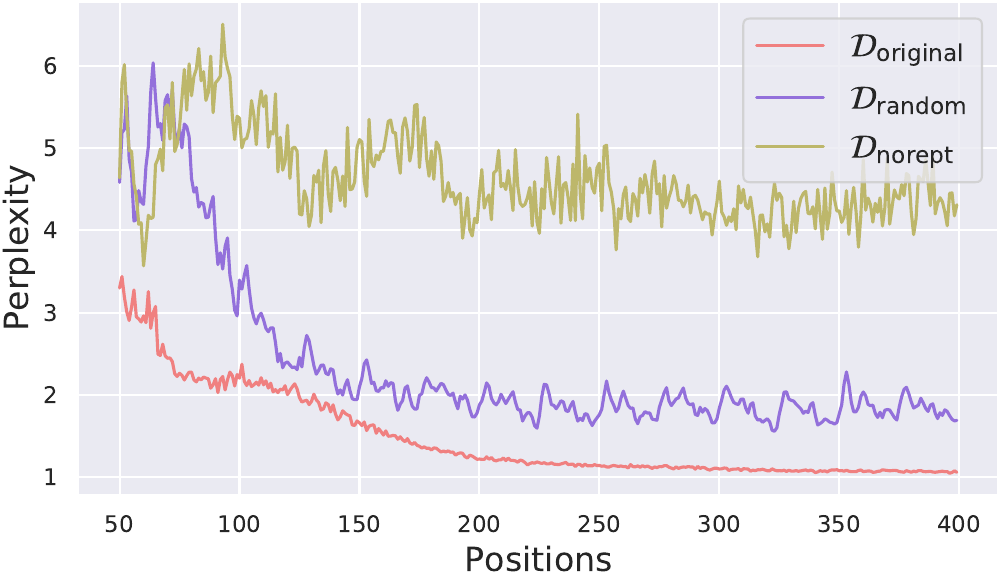}
\caption{Average per-token perplexity scores of texts generated by LMs trained on $\mathcal{D}_{\rm original}$, $\mathcal{D}_{\rm random}$ and $\mathcal{D}_{\rm norept}$ with greedy decoding. We compute their respective perplexity scores using the corresponding LMs.}
\label{ppl_self_reinforce}
\end{figure}

\paragraph{Verification} For the experiment in Table~\ref{reprep}, we use the same approach to construct the corresponding validation sets of 480 examples for $\mathcal{D}_{\rm original}$, $\mathcal{D}_{\rm random}$ and $\mathcal{D}_{\rm norept}$, and train three LMs until the best validation performance. Table~\ref{case_randomsent} shows several generation cases with greedy decoding. 
The LMs trained on $\mathcal{D}_{\rm original}$ and $\mathcal{D}_{\rm random}$ fall into repetition loops. Although the LM trained on $\mathcal{D}_{\rm norept}$ also generates sentences that have previously appeared, it does not get stuck into loops. We further investigate whether the three LMs show the self-reinforcement effect: the more times a sentence is repeated in the context, the higher the probability of continuing to generate that sentence~\cite{holtzman2019curious,xu2022learning}. Figure~\ref{ppl_self_reinforce} indicates that the LMs trained on $\mathcal{D}_{\rm original}$ and $\mathcal{D}_{\rm random}$ show the above effect, while the LM trained on $\mathcal{D}_{\rm norept}$ does not. The results suggest that longer-range repetitive patterns biased LMs to fall into repetition loops through the self-reinforcement effect whether such patterns are true or spurious. The LM trained on $\mathcal{D}_{\rm norept}$ always generate sentences in a limited set due to greedy decoding which aims to find the global maxima of probability distributions, instead of the preference for repetition loops.

\begin{table*}[!ht]
\tiny
    \centering
    \begin{tabular}{l|p{390pt}}
    \toprule 
\textbf{Training Set}&\textbf{Cases}\\
\midrule
\textbf{$\mathcal{D}_{\rm original}$}&\textbf{<eos> = Robert Boulter = <eos> <eos> Robert Boulter is an English film , television and theatre actor . He had a guest @-@ starring role on the television series The Bill in 2000 .} He has appeared in several films , including the television series The Bill , The Bill Goes to Washington , and The Bill Goes to Washington . He has also appeared in several films , including The Bill Goes to Washington , The Bill Goes to Washington , and The Bill Goes to Washington . He has also appeared in several films , including The Bill Goes to Washington , The Bill Goes to Washington , and The Bill Goes to Washington . <eos> Boulter was born in London , England , on 23 May 1986 . He is the third child of actor and actress Robert Boulter and his wife , Susan . He is the third of five children born to his wife Susan and their three children , Robert , Roberta , and Roberta . Robert Boulter 's father , Robert Boulter , was a film director and producer . He was the first actor to be cast in a film role , and the first to be cast in a television series . He was also the first actor to be cast in a television series . <eos> Boulter 's father , Robert Boulter , was a film director and producer . He was the first actor to be cast in a television series . He was the first actor to be cast in a television series . He was the first actor to be cast in a television series . He was the first actor to be cast in a television series . He was the first actor to be cast in a television series . He was the first actor to be cast in a television series . He was the first actor to be cast in a television series . He was the first actor to be cast in a television series . He was the first actor to be cast in a television series . He was the first actor to be cast in a television series . He was the first actor to be cast in a television series . He was the first actor to be cast in a television series . He was the first actor to be cast in a television series . He was the first actor to be cast in a television series . He was the first actor to be cast in a television series . He was the first actor to be cast in a television series . He was the first actor to be cast in a television series . He was the first actor to be cast in a television series\\
\midrule
\textbf{$\mathcal{D}_{\rm random}$}&\textbf{<eos> = Robert Boulter = <eos> <eos> Robert Boulter is an English film , television and theatre actor . He had a guest @-@ starring role on the television series The Bill in 2000 .} The first of these was the <unk> , a small , rectangular structure that was built in the late 19th century . The first of these was the <unk> , a small , rectangular structure that was built in the late 19th century . The first of these was the <unk> , a small , rectangular structure that was built in the late 19th century . The first of these was the <unk> , a small , rectangular structure that was built in the late 19th century . The first of these was the <unk> , a small , rectangular structure that was built in the late 19th century . The first of these was the <unk> , a small , rectangular structure that was built in the late 19th century . The first of these was the <unk> , a small , rectangular structure that was built in the late 19th century . The first of these was the <unk> , a small , rectangular structure that was built in the late 19th century . The first of these was the <unk> , a small , rectangular structure that was built in the late 19th century . The first of these was the <unk> , a small , rectangular structure that was built in the late 19th century . The first of these was the <unk> , a small , rectangular structure that was built in the late 19th century . The first of these was the <unk> , a small , rectangular structure that was built in the late 19th century . The first of these was the <unk> , a small , rectangular structure that was built in the late 19th century . The first of these was the <unk> , a small , rectangular structure that was built in the late 19th century . The first of these was the <unk> , a small , rectangular structure that was built in the late 19th century . The first of these was the <unk> , a small , rectangular structure that was built in the late 19th century . The first of these was the <unk> , a small , rectangular structure that was built in the late 19th century . The first of these was the <unk> , a small , rectangular structure that was built in the late 19th century . The first of these was the <unk> , a smal\\\midrule
\textbf{$\mathcal{D}_{\rm norept}$}&\textbf{<eos> = Robert Boulter = <eos> <eos> Robert Boulter is an English film , television and theatre actor . He had a guest @-@ starring role on the television series The Bill in 2000 .} The first two were built by the British Royal Navy . It was also released on the iTunes Store on September 28 , 2010 . It is also possible that he was a member of the royal family . He also said that he would not be returning to the team . @ 5 m ) wide and 2 feet ( 0 @. The song was written by producer and songwriter David Gilmour . It was also released on the iTunes Store on September 28 , 2010 . It was also released on the iTunes Store on September 28 , 2010 . It was also released on the iTunes Store on September 28 , 2010 . @ 5 million ( US \$ 2 @,@ 000 ) . The song was written by producer and songwriter David Gilmour . He also said that he would not be returning to the team . It was also released on the iTunes Store on September 28 , 2010 . It is also possible that he was a member of the royal family . @ 5 m ) wide and 2 feet ( 0 @. The two ships were to be joined by two smaller ships . It was also released on the iTunes Store on September 28 , 2010 . He also said that he would not be returning to the team . It was also released on the iTunes Store on September 28 , 2010 . @ 5 million ( US \$ 2 @,@ 000 ) worldwide . The song was written by David Gilmour and directed by David Gilmour . It was also released on the iTunes Store on September 28 , 2010 . It is also possible that he was a member of the royal family . He also said that he would not be returning to the team . @ 5 m ) wide and 2 feet ( 0 @. The two ships were protected by armour plates of 100 millimeters ( 3 @. It was also released on the iTunes Store on September 28 , 2010 . It was also released on the iTunes Store on September 28 , 2010 .\\
    \bottomrule
    \end{tabular}
    \caption{Cases generated by three LMs trained on different training sets with greedy decoding. The inputs are highlighted in \textbf{bold}.}
    \label{case_randomsent}
\end{table*}

\begin{table}[!ht]
\small
    \centering
    \begin{tabular}{l|cc}
    \toprule 
    \textbf{Models}&\textbf{Wikitext-103}&\textbf{WritingPrompts}\\
    \midrule
    \textbf{MLE}&0.9&0.9\\
    \textbf{UL}&0.7&0.8\\
    \textbf{ScaleGrad}&0.5&0.6\\
    \midrule
    \textbf{\textsc{SelfCont}}&0.6&0.7\\
    \bottomrule
    \end{tabular}
    \caption{Settings of $p$ for nucleus sampling.}
    \label{p_for_topp}
\end{table}

\section{Hyper-Parameters}\label{hyperparam}
We decide the hyper-parameters  $\lambda$ in Eq.~\ref{wfunc} and $p$ for nucleus sampling by searching for the value that makes the R-64 score of generated texts closest to the ground truth on the validation set. We search $\lambda$ in the range \{1.0, 2.0, 3.0, 4.0, 5.0, 6.0\}, and $p$ in the range \{0.1, 0.2, 0.3, 0.4, 0.5, 0.6, 0.7, 0.8, 0.9\}. Table~\ref{p_for_topp} shows the settings of $p$ for different models. As for baselines, we follow the original papers to set $\alpha$ to 1.0 for UL and $\gamma$ to 0.2 for ScaleGrad.

As for the choice of $f_{\theta_0}$, we empirically choose the checkpoint after training for one epoch, which allows enough training steps for self-contrastive training. We use the premature checkpoint of the same model instead of other models since different models may have different biases. It costs about 24 hours to train \textsc{SelfCont} on Wikitext-103~($\sim$10 epochs) or CNN News~($\sim$6 epochs). The results are based on one NVIDIA Tesla V100 (32GB memory) with a random single run.


\section{Modeling Token-Level Repetition}
We compare \textsc{SelfCont} with baselines in terms of the performance for modeling token-level repetition. As shown in Table~\ref{rep_prediction}, \textsc{SelfCont} achieves  higher overall accuracy, higher F1 score on non-repetitive tokens, and comparable F1 score on repetitive tokens.

\begin{table}[!ht]
\scriptsize
    \centering
    \begin{tabular}{l|c|ccc|ccc}
    \toprule 
    \multirow{2}{*}{Models}&\multirow{2}{*}{\textbf{Acc}}&\multicolumn{3}{c|}{\textbf{Repetitive}}&\multicolumn{3}{c}{\textbf{Non-Repetitive}}\\
    &&\textbf{P}&\textbf{R}&\textbf{F1}&\textbf{P}&\textbf{R}&\textbf{F1}\\
    \midrule
    \textbf{MLE}&78.9&67.4&\textbf{87.0}&75.9&\textbf{90.2}&73.9&81.2\\
    \textbf{UL}&80.5&70.9&\underline{83.0}&\textbf{76.5}&\underline{88.2}&78.9&83.3\\
    \textbf{ScaleGrad}&\underline{81.6}&\underline{75.9}&76.1&76.0&85.2&\underline{85.0}&\underline{85.1}\\
    \midrule    \textbf{\textsc{SelfCont}}&\textbf{82.3}&\textbf{78.5}&73.9&\underline{76.1}&84.4&\textbf{87.4}&\textbf{85.9}\\

    \bottomrule
    \end{tabular}
    \caption{Repetition prediction performance on the test set of Wikitext-103. \textbf{Acc} is the overall accuracy of predicting whether the next token is repetitive. We regard the model-predicted token (with the maximum probability) as repetitive if it has appeared in the previous 64 tokens (non-repetitive otherwise).  \textbf{P}, \textbf{R} and \textbf{F1} means the precision, recall and F1 on the class of  (non-)repetitive tokens.}
    \label{rep_prediction}
\end{table}

\section{Case Study}
Table~\ref{case_model_baseline} and Table~\ref{case_model_baseline_topp} show the cases generated by different models on Wikitext-103 with greedy decoding and nucleus decoding, respectively. We see that \textsc{SelfCont} can still get stuck into loops with greedy decoding since it hardly learns longer-range dependencies than standard LMs. Although sampling helps reduce superficial repetition, it does not utilize underlying long-range dependencies to maintain long-range coherence. Therefore, it is important to improve the modeling of long-range dependencies to essentially solve the repetition problem in future work.


\begin{table*}[!ht]
\tiny
    \centering
    \begin{tabular}{l|p{390pt}}
    \toprule 
\textbf{Models}&\textbf{Cases (Greedy Decoding)}\\
\midrule
\textbf{MLE}&\textbf{<eos> = Robert Boulter = <eos> <eos> Robert Boulter is an English film , television and theatre actor . He had a guest @-@ starring role on the television series The Bill in 2000 .} He has also appeared in the television series The Bill , The Bill 2 , The Bill and The Bill 3 . <eos> Boulter was born in London , England , and grew up in the United States . He studied at the University of California , Berkeley , and the University of California , Berkeley . He was a member of the British Academy of Film and Television Arts ( BAFTA ) in 2001 . He has appeared in several television series , including The Bill , The Bill 2 , The Bill 3 , The Bill 4 , The Bill 5 , The Bill 6 , The Bill 7 , The Bill 8 , The Bill 9 , The Bill 10 , The Bill 11 , The Bill 12 , The Bill 13 , The Bill 14 , The Bill 15 , The Bill 16 , The Bill 17 , The Bill 18 , The Bill 19 , The Bill 20 , The Bill 21 , The Bill 22 , The Bill 23 , The Bill 24 , The Bill 25 , The Bill 26 , The Bill 27 , The Bill 28 , The Bill 29 , The Bill 30 , The Bill 31 , The Bill 32 , The Bill 33 , The Bill 34 , The Bill 34 , The Bill 35 , The Bill 36 , The Bill 37 , The Bill 39 , The Bill 40 , The Bill 41 , The Bill 42 , The Bill 43 , The Bill 44 , The Bill 45 , The Bill 46 , The Bill 47 , The Bill 48 , The Bill 49 , The Bill 50 , The Bill 51 , The Bill 52 , The Bill 53 , The Bill 54 , The Bill 55 , The Bill 56 , The Bill 57 , The Bill 58 , The Bill 59 , The Bill 60 , The Bill 61 , The Bill 62 , The Bill 63 , The Bill 64 , The Bill 65 , The Bill 66 , The Bill 67 , The Bill 68 , The Bill 69 , The Bill 70 , The Bill 71 , The Bill 72 , The Bill 73 , The Bill 74 , The Bill 75 , The Bill 76 , The Bill 77 , The Bill 78 , The Bill 79 , The Bill 80 , The Bill 81 , The Bill 82 , The Bill 83 , The Bill 84 , The Bill 85 , The Bill 86 , The Bill 87 , The Bill 88 , The Bill 89 , The Bill 90 , The Bill 91 , The Bill 92 , The Bill 93\\
\midrule
\textbf{UL}&\textbf{<eos> = Robert Boulter = <eos> <eos> Robert Boulter is an English film , television and theatre actor . He had a guest @-@ starring role on the television series The Bill in 2000 .} He has also appeared in several films , including the comedy @-@ drama The Man Who Would Be King ( 2002 ) , the science fiction film The Man Who Would Be Super ( 2004 ) , the science fiction film The Man Who Would Be Super ( 2006 ) , the science fiction film The Man Who Would Be Super ( 2008 ) , the science fiction film The Man Who Would Be Super ( 2012 ) , the science fiction film The Day After Tomorrow ( 2013 ) , the science fiction film The Day After Tomorrow ( 2014 ) , the science fiction film The Day After Tomorrow ( 2015 ) , the science fiction film The Day After Tomorrow ( 2016 ) , the science fiction film The Day After Tomorrow ( 2017 ) , the science fiction film The Day After Tomorrow ( 2017 ) , the science fiction film The Day After Tomorrow ( 2017 ) , the science fiction film The Day After Tomorrow ( 2017 ) , the science fiction film The Day After Tomorrow ( 2017 ) , the science fiction film The Day After Tomorrow ( 2017 ) , the science fiction film The Day After Tomorrow ( 2017 ) , the science fiction film The Day After Tomorrow ( 2017 ) , the science fiction film The Day After Tomorrow ( 2017 ) , the science fiction film The Day After Tomorrow ( 2017 ) , the science fiction film The Day After Tomorrow ( 2017 ) , the science fiction film The Day After Tomorrow ( 2017 ) , the science fiction film The Day After Tomorrow ( 2017 ) , the science fiction film The Day After Tomorrow ( 2017 ) , the science fiction film The Day After Tomorrow ( 2017 ) , the science fiction film The Day After Tomorrow ( 2017 ) , the science fiction film The Day After Tomorrow ( 2017 ) , the science fiction film The Day After Tomorrow ( 2017 ) , the science fiction film The Day After Tomorrow ( 2017 ) , the science fiction film The Day After Tomorrow ( 2017 ) , the science fiction film The Day After Tomorrow ( 2017 ) , the science fiction film The Day After Tomorrow ( 2017 ) , the science fiction film The Day After Tomorrow ( 2017 ) , the science fiction film The Day After Tomorrow ( 2017 ) , the science fiction film The Day After Tomorrow ( 2017 ) , the science fiction film The Day After Tomorrow ( 2017 ) , the science fiction film The Day After Tomorrow ( 2017 ) , the science fiction film The Day After Tomorrow\\
\midrule
\textbf{ScaleGrad}&\textbf{<eos> = Robert Boulter = <eos> <eos> Robert Boulter is an English film , television and theatre actor . He had a guest @-@ starring role on the television series The Bill in 2000 .} In 2002 he appeared as a character in the BBC 's crime drama series The Secret Service . He has also worked as a consultant for several films including The Man Who Would Be King ( 2004 ) , The Man Who Would Never Die ( 2007 ) , The Man Who Would Never Be King 2 ( 2009 ) , The Man Who Would Never Be King 3 ( 2011 ) , The Man Who Would Never Be King 4 ( 2013 ) , The Man Who Would Never Be King 5 ( 2014 ) , The Man Who Would Never Be King 6 ( 2015 ) , The Man Who Would Never Be King 7 ( 2016 ) , The Man Who Would Never Be King 8 ( 2017 ) , The Man Who Would Never Be King 9 ( 2017 ) , The Man Who Would Never Be King 10 ( 2017 ) , The Man Who Would Never Be King 11 ( 2017 ) , The Man Who Would Never Be King 12 ( 2017 ) , The Man Who Would Never Be King 13 ( 2017 ) , The Man Who Would Never Be King 14 ( 2017 ) , The Man Who Would Never Be King 15 ( 2017 ) , The Man Who Would Never Be King 16 ( 2017 ) , The Man Who Would Never Be King 17 ( 2017 ) , The Man Who Would Never Be King 18 ( 2017 ) , The Man Who Would Never Be King 19 ( 2017 ) , The Man Who Would Never Be King 20 ( 2017 ) , The Man Who Would Never Be King 21 ( 2017 ) , The Man Who Would Never Be King 22 ( 2017 ) , The Man Who Would Never Be King 23 ( 2017 ) , The Man Who Would Never Be King 24 ( 2017 ) , The Man Who Would Never Be King 25 ( 2017 ) , The Man Who Would Never Be King 26 ( 2017 ) , The Man Who Would Never Be King 27 ( 2017 ) , The Man Who Would Never Be King 28 ( 2017 ) , The Man Who Would Never Be King 29 ( 2017 ) , The Man Who Would Never Be King 30 ( 2017 ) , The Man Who Would Never Be King 31 ( 2017 ) , The Man Who Would Never Be King 32 ( 2017 ) , The Man Who Would Never Be King 33 ( 2017 ) , The Man Who Would Never Be King 34 ( 2017 ) , The Man Who Would Never Be King 35 ( 2017 ) , The Man Who Woul\\
\midrule
\textbf{\textsc{SelfCont}}&\textbf{<eos> = Robert Boulter = <eos> <eos> Robert Boulter is an English film , television and theatre actor . He had a guest @-@ starring role on the television series The Bill in 2000 .} Boulter appeared in two episodes of the British comedy @-@ drama Oz in 2001 , playing an American businessman who helps an Italian woman ( Joanna Cassidy ) escape from her abusive husband ( Michael Sheen ) . He also played a character in the 2004 film Heat , which was directed by Michael Medvedev . Boulter appeared in two episodes of the British comedy @-@ drama Body of Proof in 2005 , playing a retired police officer who is accused of murdering his wife ( Christine Fay ) . He appeared in two episodes of the British television drama Body of Proof in 2006 , playing a retired judge who defends his client , Simon Ashdown ( Stephen Fry ) . Boulter also appeared in the Australian television series The X Factor in 2007 , playing a retired judge who defends his client , Kiernan Straker ( Simon Ashdown ) . <eos> Boulter was nominated for a Golden Globe Award for Best Actor – Motion Picture Musical or Comedy , but lost to Hugh Grant for the role of Seamus Heaney . In 2010 , he appeared in the British comedy @-@ drama The X Factor , playing a retired judge who is accused of murdering his wife ( Christine Fay ) . He also appeared in the Australian television series The X Factor in 2010 , playing a retired judge who defends his client , Simon Ashdown . In 2012 , he appeared in the British comedy @-@ drama Body of Proof , playing a retired judge who is accused of murdering his wife ( Joanna Cassidy ) . He also appeared in the American television series The X Factor in 2013 , playing a retired judge who is accused of murdering his wife ( Christine Fay ) . <eos> Boulter has appeared in a number of films , including the 2005 film Heat , and the 2008 comedy Body of Proof . He appeared in the British television series Heat 2 , and the 2009 comedy Body of Proof . In 2011 , he appeared in the Australian television series The X Factor . In 2012 , he appeared in the British television series The X Factor . In 2013 , he appeared in the British television series The X Factor . In 2014 , he appeared in the Australian television series The X Factor . In 2015 , he appeared in the British television series The X Factor . In 2016 , he appeared in the Australian television series The X Factor\\
\bottomrule
    \end{tabular}
    \caption{Cases generated by different models with greedy decoding on Wikitext-103. The inputs are highlighted in \textbf{bold}.}
    \label{case_model_baseline}
\end{table*}

\begin{table*}[!ht]
\tiny
    \centering
    \begin{tabular}{l|p{390pt}}
    \toprule 
\textbf{Models}&\textbf{Cases (Nucleus Decoding)}\\
\midrule
\textbf{MLE}&\textbf{<eos> = Robert Boulter = <eos> <eos> Robert Boulter is an English film , television and theatre actor . He had a guest @-@ starring role on the television series The Bill in 2000 .} Since 1994 , he has appeared in a wide range of productions . He has played Crocodile Hunter , a domestic terrorist in A Change Is Gonna Come . \ul{His first feature film role was in a 1983 film , The Perfect Numbers , written by Ridley Scott .} A science fiction parody of the British mafia , his next film role was the horror comedy The Imperfect People . The film was a box office hit and has been named one of the greatest movies ever made . <eos> Boulter portrayed a divorcee living in Chicago , Illinois , from the start of its development in 1986 , when he met Hollywood agent Kevin De Matos . The two became very close and married on August 29 , 1988 . He has become a great fan of James Cameron 's 1999 film Avatar . In December 2011 , he starred as a roadie who stands in a line at a restaurant . <eos> <eos> = = Early life = = <eos> <eos> Robert Boulter was born in Dundee , Scotland , on November 6 , 1961 , and raised in Dundee , Scotland . He attended Eales College , Dundee . He received an MBE for his work at Eales as a schoolteacher in 1973 , and graduated from Dundee University in 1974 . After teaching history to three young teenagers , he acted in numerous short films . <eos> <eos> = = Career = = <eos> <eos> \underline{He made his acting debut in 1976 with the short film The Quarryman , playing the character Andre Taylor in a variety of movies and television shows .} Boulter has appeared on television and cinema advertisements as well . <eos> In 1982 , Boulter co @-@ starred in the video game Dr. No . \ul{The following year he made his film debut in the video game E1 Story , played by Terry Gilliam .} He also appeared in the 1984 video game Doctor Who , starring Matthew Broderick . <eos> Boulter met producer Shane Bobbitt in 1982 , and the two became friends , and Broderick made Boulter his stand @-@ up comedy partner . On the strength of their relationship , Boulter starred in several feature film projects for the American television\\
\midrule
\textbf{UL}&\textbf{<eos> = Robert Boulter = <eos> <eos> Robert Boulter is an English film , television and theatre actor . He had a guest @-@ starring role on the television series The Bill in 2000 .} He starred as the title character in 2008 's A Charlie Brown Christmas , directed by Adam Mickiewicz . Boulter has also played a variety of other roles over the years , including as the titular character on <unk> USA in 2002 , in the <unk> anthology series Blood 's Wild in 2007 , as the protagonist on the murder mystery A Beautiful Guy in 2009 , and as the love interest for the titular character on Queen 's College in 2012 . Boulter has also played the supporting roles of Richard VIII and Queen Elizabeth in various media including television , film , and video games . He won the Academy Award for Best Actor for his performance in King George V. <eos> \ul{Born in Carlisle , Lancashire , Boulter grew up in Manchester .} He has been involved in theatre since he was eight years old . His only film role came in 2000 , when he played Jack Leppard in Harry Potter and the Goblet of Fire . A second appearance came in 2001 , when he played the young Severus Snape in Romeo and Juliet . He has worked on television shows including , Steven Soderbergh 's From Russia , with Love , Watchmen , Dawson 's Creek , Ealing Studios and the British Channel 7 documentary series Ghost Hunters . <eos> Boulter has made four appearances on television : seven as Colin Rudge in Star Trek : Voyager ( 1995 – 1997 ) ; 13 as Ian Frazier in The Sarah Jane Adventures ( 1997 – 1999 ) ; and 16 as Scott Reid in Michael Bay 's Robin Hood : How the Legend Was Won ( 1999 ) . He was nominated for a Laurence Olivier Award in 2001 . <eos> <eos> = = Early life = = <eos> <eos> \ul{Boulter was born on 12 April 1979 in Carlisle , Lancashire , England .} His father is a retired pilot . He attended Elgin Grammar School , where he earned an academic degree . After leaving school , he worked in retail at an engineering firm in Manchester . <eos> Boulter attended <unk> High School , Carlisle before transferring to King 's College , Cambridge . During this time , he became involved in theatre and became a student at Queen 's College , Cambridge .\\
\midrule
\textbf{ScaleGrad}&\textbf{<eos> = Robert Boulter = <eos> <eos> Robert Boulter is an English film , television and theatre actor . He had a guest @-@ starring role on the television series The Bill in 2000 .} \ul{In 2004 he starred as the character Sherlock Holmes in the BBC medical drama Holby City .} \ul{Boulter made his screen debut with the comedy series Chitty Chitty Bang Bang , which was broadcast in 2006 .} He then played Lord Stanley in The Lord of the Rings trilogy and in Pirates of the Caribbean : At World 's End . In 2007 he portrayed David Marmion in the film adaptation of A Midsummer Night 's Dream . <eos> He appeared in the British film Downton Abbey ( 2008 ) , which won him a Golden Globe Award for Best Actor . He also starred in Steven Spielberg 's fantasy film , The Wrestler , as the title character . <eos> In 2010 , Boulter portrayed Leonardo DiCaprio in the directorial debut of Woody Allen 's The Wrestler . The film tells the story of two men who live in Los Angeles , California , who have been feuding over money and power . The film received mixed reviews from critics . <eos> <eos> = = Early life = = <eos> <eos> Boulter was born in Leicester , England , on 3 October 1963 . His father , John , is a chemical engineer and designer . His mother , Mary ( née <unk> ) , is a member of the Scottish Episcopal Church . He grew up in Loughborough , Leicester . He was educated at Eton College and then St Thomas 's School , Northamptonshire . His older brother , Liam , plays football for Leicester City . <eos> Boulter started playing football when he was four years old . After a few years , he joined Brentford Town , where he played alongside Ronnie Brown . Boulter enjoyed the game and liked the fact that his father had taught him how to play football . When he turned twelve , he moved to Leicester City , but he left the club after one season because of disciplinary problems . Boulter did not attend Brentford 's youth team , but took up playing football for them . During his time at Brentford , he played for several clubs including Chesterfield , Rochdale , Oldham Athletic , St James ' Park , Scarborough and Lewes . He made his debut for the club aged 15 in 1971 , an\\
\midrule
\textbf{\textsc{SelfCont}}&\textbf{<eos> = Robert Boulter = <eos> <eos> Robert Boulter is an English film , television and theatre actor . He had a guest @-@ starring role on the television series The Bill in 2000 . }He also appeared in a 2000 episode of Syfy 's Geneva Live . <eos> Boulter is an accomplished box @-@ office actor and one of the best known box @-@ office draws in the history of British television . Boulter appeared in over 20 films and received many awards , including four Academy Awards , including Best Actor and Best Supporting Actor , and the BAFTA Award for Best British Actor . He was nominated for six other BAFTAs , winning three , for his work on the television series and the 1997 film . He starred in The Bill in 2001 and again in 2002 . In 2005 , he appeared in The Gleason Room , the 2005 science fiction film about rediscovery of woolly alien relics , and in the 2006 biographical drama Brand New Eyes . In 2010 , he starred in the stage production of Minor Threat and the 2007 psychological thriller Victoria 's Secret . <eos> Boulter 's stage and film career began with his performance in the 1997 romantic comedy Hamlet . In 2000 , he was cast as Jonathan Simeone in the German @-@ language dramatisation of French novelist Raymond Lebowski 's epic play , The Professionals . He took on the role of " Troy " , an obsessive person who attempts to prove himself to a courtiers . Although he enjoyed playing Troy , he took " enormous risks " , in the words of theatre critic Graham McCann , who wrote that " there was nothing to lose in playing a man like Troy . " He co @-@ starred in The Professionals with Julianne Moore and Kim Novak . He portrayed the criminal Tammi Martineau in the 2004 biographical film Asterisk and appeared in several films and television shows . In 2005 , he starred as Garth Snow in the Fox crime drama Dangerous Liaisons . <eos> Boulter is known for his film work in Hungary and abroad . He has also worked with Brandon Thomas and Sacha Baron Cohen . In 2011 , he was nominated for a Laurence Olivier Award for Best Actor , with Olivier in the role of General Herculaneum . In 2012 , he starred in The Phantom of the Opera , which opened at the BBC2 Leicester Square Theatre , with much of the stage cast from his earlier work\\
\bottomrule
    \end{tabular}
    \caption{Cases generated by different models with nucleus decoding on Wikitext-103. The inputs are highlighted in \textbf{bold}, while the incoherent sentences are \underline{underlined}.}
    \label{case_model_baseline_topp}
\end{table*}
\end{document}